**What Do Large Language Models Know? Tacit Knowledge as a Potential Causal-Explanatory Structure**

Céline Budding[1]

[1] Philosophy & Ethics group and Eindhoven Artificial Intelligence Systems Institute, Eindhoven University of Technology, c.e.budding@tue.nl

**Abstract**

It is sometimes assumed that Large Language Models (LLMs) know language, or for example that they know that Paris is the capital of France. But what—if anything—do LLMs actually know? In this paper, I argue that LLMs can acquire *tacit knowledge* as defined by Martin Davies (1990). Whereas Davies himself denies that neural networks can acquire tacit knowledge, I demonstrate that certain architectural features of LLMs satisfy the constraints of semantic description, syntactic structure, and causal systematicity. Thus, tacit knowledge may serve as a conceptual framework for describing, explaining, and intervening on LLMs and their behavior.

# 1. Introduction

Although large language models (LLMs) are generally trained on next-word prediction tasks, these systems have exhibited impressive performance in the generation of seemingly human-like, coherent texts (e.g. ChatGPT (OpenAI 2022)). Yet, we do not know how these models work and what they learn from their data. Like other types of deep neural networks, LLMs are opaque and suffer from the black-box problem: during training, LLMs learn a highly complex function with distributed representations and we cannot simply "look inside" to determine how they work (Burrell 2016; Creel 2020). Because of this, it is difficult to evaluate their potential linguistic and cognitive capacities (Shanahan 2023).

The field of *explainable AI* (XAI) aims to solve the problem of opacity by developing explanations through mathematical techniques that show how or why a network makes a certain decision (Zednik 2021). One specific question that has been asked is whether LLMs learn any kind of rules or algorithms, akin to symbolic rules in classical systems (Olah et al. 2020; Pavlick 2023). Uncovering such rules would help explain how AI systems work, facilitate the evaluation of potential cognitive and linguistic capacities, and potentially allow for interventions in the model internals to update the system's predictions.

In the particular case of LLMs, it has been suggested that they do not just perform next-word prediction based on surface statistics, but that they learn to follow symbolic rules (Pavlick 2023), develop linguistic competence (Mahowald et al. 2024), or even acquire a kind of knowledge (Meng et al. 2022; Yildirim and Paul 2024). Indeed, with respect to syntax, LLMs seem to successfully represent grammatical rules, and apply formal linguistic structure (Linzen and Baroni 2021; Mahowald et al. 2024). Less attention has been paid to semantics, but some very recent studies claim to have identified representations of world models (Li et al. 2023), implicit meaning (Hase et al. 2021), and facts (De Cao, Aziz, and Titov 2021; Meng et al. 2022). Although it is clear that many of these previous discussions are concerned with what LLMs "know", it remains difficult to conceptualize exactly what this "knowledge" actually consists in. In this paper, I will focus on semantics, and in particular, LLMs' knowledge of semantic rules considered as representations of facts or other propositions that play a causal role in the system's behavior.

Specifically, taking inspiration from debates between symbolic and connectionist AI in the 1980s and 90s (e.g. Clark 1991; Fodor and Pylyshyn 1988), I propose that *tacit knowledge*, as defined by Martin Davies (1990), provides a suitable way to conceptualize semantic knowledge in LLMs. Tacit knowledge, in this context, refers to implicitly represented rules that causally affect the system's behavior. As connectionist systems are known not to have explicit knowledge, tacit knowledge provides a promising way to characterize and identify meaningful representations in the model internals. Moreover, if representations of knowledge can be successfully identified, these could be used both to explain how the model works and to potentially change the behavior of the system by overwriting or otherwise changing this knowledge.

This paper is not the first to propose tacit knowledge for connectionist neural networks. As mentioned previously, Davies (1990) originally developed the account of tacit knowledge as a way to attribute knowledge to connectionist systems in the absence of explicit rules. Nevertheless, Davies argues that connectionist neural networks do not actually meet the constraints needed for attribution of tacit knowledge. As such, my contribution goes beyond Davies by applying tacit knowledge to contemporary transformer-based LLMs and showing that these can in fact meet the constraints. More recently, Lam (2022) has argued that the identification and description of tacit knowledge should be a target for explainable AI, but does not demonstrate that neural networks can actually have such tacit knowledge.

    Overall, I claim that if LLMs meet Davies' constraints, they are capable of acquiring a form of tacit knowledge, and that there is preliminary evidence that some LLMs have in fact done so. To support this argument, I will 1) discuss why LLMs, while they are trained to perform next-word prediction, might actually learn semantic rules, 2) introduce Davies' account of tacit knowledge and the constraints that should be met to attribute such knowledge, 3) address Davies' objection to applying tacit knowledge to more complex connectionist networks and show this does not hold for LLMs, and 4) analyze an example from the recent technical literature to demonstrate that some current LLMs might actually meet these constraints. Taken together, the contributions of this paper are twofold. First, I suggest that tacit knowledge, as defined by Davies (1990) is a promising way to conceptualize semantic knowledge in LLMs and show how the constraints can be modified to apply to current systems. Second, I show that there is convincing—albeit preliminary—evidence that at least some LLMs meet these constraints.

## 2. A Short Introduction to Large Language Models

### 2.1 The transformer architecture

Although LLMs have recently gained immense popularity, creating a computational (rule-based or probabilistic) model of language has been the goal of natural language processing (NLP) since the 1950s, with machine learning-based methods being available since the 1980s (Nadkarni, Ohno-Machado, and Chapman 2011). Like other machine learning-based neural networks, LLMs are trained on a particular task, generally a *next-word prediction* task.[1] In the simplest version of this task, a network receives a particular input sequence and needs to predict the most likely next word based on a learned probability distribution over the vocabulary. User interfaces like ChatGPT perform this task repeatedly to produce individual sentences or even essay-length texts.

The difficulty of this task lies in the fact that the next word(s) in a sequence often depend on the context given by all previous text, and not just the words directly preceding it. While earlier NLP systems were fairly successful overall, they generally struggled with long-range dependencies (Graves 2012), leading to incoherent or repeating output text. Current LLMs, based on the *transformer* architecture (Vaswani et al. 2017), do better at integrating context, even when generating long pieces of text. What makes the transformer architecture particularly suitable for language is the addition of an *attention mechanism* (Bahdanau, Cho, and Bengio 2015)*,* which allows the network to keep track of long-range dependencies and integrate information about previous words (i.e. context) in its predictions. That is, the attention mechanism allows the network to learn which previous words or parts of the input sequence it should focus on when processing a particular word.

---

[1] Strictly speaking, LLMs predict *tokens* rather than *words*. Although tokens can and do sometimes correspond to words, they can also refer to other subsets of the input sequence, such as parts of words, single letters, or punctuation. For sake of simplicity, however, I will talk about words, even when I should technically speak about tokens.

      More specifically, transformer models consist of an input layer, an embedding layer (see section 4), an output layer that returns the final prediction, and stacked transformer blocks in between (figure 1). These transformer blocks themselves consist of two layers: the attention layer and a fully-connected multilayer perceptron (MLP) layer. While MLP layers have not been discussed as much as the attention layer, they have recently been suggested to play a role in enriching the internal representations with relevant context (Geva et al. 2021; Sukhbaatar et al. 2019, also see section 5). After the model has been trained by tuning its weights for a particular task, the network can predict the most probable next word given the probability distribution over the vocabulary and the context given by the input sequence itself.

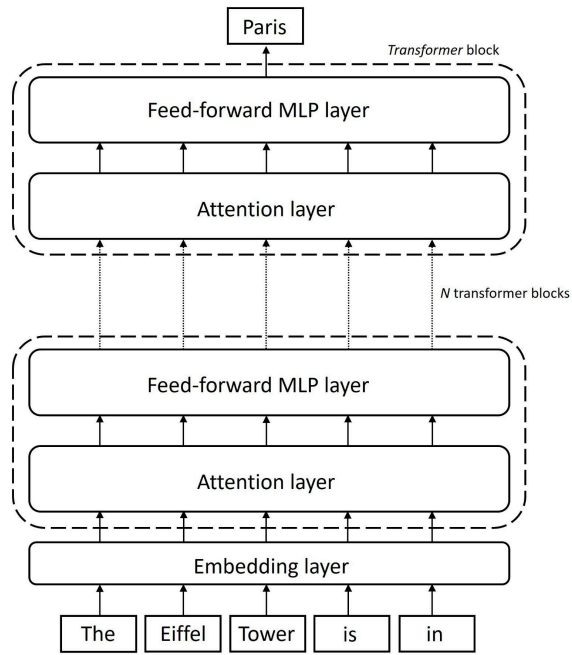

Figure 1: Simplified overview of the transformer architecture. At the bottom, the input is "The Eiffel Tower is in". The embedding layer computes word embeddings for each word in the input (section 4). Afterwards, the embedding is processed by a number of stacked transformer blocks, each consisting of an attention and MLP layer (section 5). Finally, at the top, the most likely word, in this case "Paris", is predicted as the output.

## 2.2 What do LLMs learn?

Although LLMs are optimized for next-word prediction, their internal processing remains opaque. During the training process, LLMs learn a highly nonlinear function that allows them to predict the most likely next word. In an ideal case, this function tracks relevant features of the data, such as correlations between words in language tasks. Nevertheless, the system might also learn to track spurious correlations—correlations in the dataset that may be considered irrelevant for a particular prediction. An example of this was identified by Lapuschkin and colleagues (2019) in a well-performing vision model. Despite its performance, they found that this model learned so-called "Clever Hans" solutions for some input-output pairs. For example, although images containing a horse were correctly classified as "horse", this classification was based on the presence of a watermark in the input, rather than on the depiction of a horse. Upon closer inspection of the dataset, it turned out that all images of horses in the training set contained such a watermark, leading the model to track this spurious correlation. As the data on which the model was tested were drawn from the same dataset, this problem had not been identified before, but could have led to poor performance in other contexts.

While the example above might seem innocent, the fact that it is unclear which rules or features neural networks learn from the data is problematic for multiple reasons. First, it might be that the system learns incorrect or biased rules or features. For instance, it might be that while learning the typical characteristics of medical doctors—e.g., that they wear white coats and carry stethoscopes—the system learns that being male is one such characteristic, possibly on the basis of oft-repeated stereotypes. Knowing which information is used by LLMs for their predictions—and ideally, where this information is stored—might facilitate interventions that can help fix the predictions of these systems.

Second, and more important for the purposes of this paper, the impressive performance of LLMs has led to the reflex-like attribution of linguistic and cognitive capacities like understanding and knowledge (e.g. Piantadosi and Hill 2022; Yildirim and Paul 2024). It is unclear and even contested, however, whether such attributions are warranted and what it means to attribute something like knowledge to an LLM. Investigating the underlying processing of these systems might provide clarity on whether the attribution of such capacities to LLMs is actually warranted (Buckner 2023; Zednik 2018).

**2.3 Attributing knowledge to LLMs**

Studying the internal processing of LLMs might help determine whether LLMs merely use statistical correlations in the text to predict the next word, or if they acquire something like knowledge, which might then explain their performance. There is fairly robust evidence that LLMs learn and represent grammatical rules, i.e. syntax (Linzen and Baroni 2021; Mahowald et al. 2024). In contrast, technical research on semantic aspects of knowledge, such as representations of world models (Li et al. 2023; Yildirim and Paul 2024) or facts (Geva et al. 2021; Meng et al. 2022), is still in early stages.

Recent work on identifying representations of semantic facts in LLMs reports promising results, however. For example, Geva and colleagues (2021) show that LLMs acquire so-called key-value memories, which seem to correlate with human-interpretable patterns such as TV series, time ranges, and specific words like 'press'. Similarly, Meng and colleagues (2022) identify representations of facts in the model internals, which they successfully edit to change the output of the network, both for the target input-output pair as well as semantically similar input-output pairs. For example, assuming that a network represents "The Eiffel Tower is located in Paris", they update this factual association to "The Eiffel Tower is located in Berlin". Afterwards, the network responds to subsequent prompts as if the Eiffel Tower is now located in Berlin, for example stating that one passes the Brandenburg Gate when walking to the Eiffel Tower (see section 5). These results are in line with other recent studies (De Cao, Aziz, and Titov 2021; Geva et al. 2021) that identify and edit representations of semantic facts in the model internals.

Nevertheless, this literature is largely motivated by practical concerns such as facilitating edits, also referred to as interventions, to improve the performance of AI systems, or explaining how the system works to identify when a system might make mistakes. As such, there is limited consideration of what it means to attribute knowledge in the first place. Indeed, there is no clear definition of what kind of knowledge LLMs might acquire, nor of the conditions that should be met in order to attribute such knowledge. In this paper, I propose one way to conceptualize this kind of knowledge in LLMs: tacit knowledge as defined by Davies (1990).

**3. Davies' Account of Tacit Knowledge**

**3.1 Knowledge in neural networks**

Following Davies, I will motivate the focus on tacit knowledge specifically by first introducing two alternative notions of knowledge of rules: explicit representation as a strong notion of knowledge, and mere conformity as a weak notion of knowledge. Symbolic AI systems such as expert systems like *Cyc* (Lenat and Guha 1989), are often said to have *explicit knowledge* of rules, in the sense that they have a pre-programmed knowledge base containing representations of rules that are invoked during processing.

These rules might be explicitly represented in the system's source code, in which case an external observer might be able to "read off" the rules given sufficient background knowledge, access, and technical expertise. Alternatively, a system might make use of a database containing rules or logical statements. While this database might not be accessible to an external observer, any rules or statements in the database can be considered explicit knowledge to the system.

Explicit knowledge plays a causal role in mediating transitions from inputs to outputs. When processing a particular input, the system needs to identify and retrieve the appropriate piece of knowledge, which then plays a causal role in the transition to a particular output. For example, in a question-answering system, this knowledge could be a statement of the form "Paris is the capital of France". When the model is queried with an input "What is the capital of France?", it returns "Paris" because of this explicitly represented knowledge. Thus, given that a system is known to have explicit knowledge, it is possible to explain its behavior in virtue of the system having this knowledge.

Connectionist systems like neural networks, on the other hand, cannot be said to have explicit knowledge of rules (see also Fodor and Pylyshyn 1988). In contrast to symbolic systems, neural networks do not invoke rules represented in a knowledge base or database while processing inputs. Instead, neural networks are optimized using machine learning techniques to perform a particular function. This means that the way an input is processed is determined by the weights of the network, rather than by explicit rules stored elsewhere in the network. As the weights of individual nodes generally do not correspond one-to-one to rules due to distributed representation (Hinton, McClelland, and Rumelhart 1986; Van Gelder 1992), neural networks cannot be said to have knowledge of rules in the strong sense.

Instead, neural networks can be said to conform to rules, or have weak knowledge of rules (Davies 1990). Knowledge of rules in this sense does not require the system to represent rules in any way, but only requires that the behavior of the system—in terms of input-output transitions—matches what is described by the rule in question. Imagine we have a network that can correctly classify images of cats and dogs. Given that it successfully performs this task, the network could be said to conform to a rule "pointy ears → cat" and "floppy ears → dog", but it could also and alternatively be said to conform to the rule "slitted pupils → cat" and "round pupils → dog". This kind of indeterminacy shows the limitation of only attributing knowledge of rules in the weak sense: one network might conform to multiple sets of rules (Kripke 1982), or multiple networks might conform to the same sets of rules while relying on wildly different underlying processing mechanisms (see e.g. Block 1981). To distinguish between these kinds of systems, an intermediate notion of knowledge is needed that characterizes knowledge of rules without requiring explicit representation.

**3.2 Tacit knowledge as an intermediate notion of knowledge**

Davies' account of tacit knowledge is meant to provide an intermediate account of knowledge, one that is more than mere conformity but less than explicit knowledge. Tacit knowledge, according to this account, refers to rules that are not represented explicitly, but that nevertheless describe causally-relevant structures that guide behavior (Davies 1990).

Tacit knowledge is different from explicit knowledge in the sense that an observer (e.g. computer programmer) cannot straightforwardly "read off" the rules by inspecting the system, but rather requires significant interpretative labor. While it might be possible for such an observer to determine whether or not a classical system has knowledge of a particular rule by inspecting its source code or knowledge base, identifying tacit knowledge of a rule—for instance in an LLM or another artificial neural network—requires interpretation. As will be discussed in section 5, this kind of interpretation might be facilitated through the use of XAI methods which determine what is represented by particular structures within the system and how these structures contribute to the system's behavior. As such, tacit knowledge differs from explicit knowledge in the amount of epistemic or interpretative labor that is needed to identify it.

In addition to this epistemic way of distinguishing between tacit and explicit knowledge, it might also be possible to draw a metaphysical distinction. In this case, the relevant distinguishing factor might be the way in which a rule can—in principle—be encoded in different architectures. To illustrate this difference, consider the case of a network implementing a syntactic rule, for example regarding subject-verb agreement. If a neural network learns this rule, it might be possible to identify where it is implemented through the use of XAI. In this case, the system would have tacit knowledge in an epistemic sense, as additional tools would be necessary for identifying the encoded rules. Yet, this system would have explicit knowledge in a metaphysical sense, insofar as it is in principle possible to describe and interpret the encoded rule. In contrast, a system would have tacit knowledge in the metaphysical sense when the causally-relevant structures in the internal processing cannot be given a semantic description even in principle, despite the fact that these structures are actually present and guide the system's behavior.[2] That said, even on this metaphysical conception of tacit knowledge, it remains an empirical question whether, and how, such causally-relevant structures can actually be identified, which is the question I will focus on in the remainder of the paper.

Tacit knowledge is also different from mere conformity to rules, insofar as it offers a potential causal explanation of the model's observed behavior. An attribution of tacit knowledge is grounded in a particular internal causal-explanatory structure (see section 3.3.2.), which can for example be used to predict the effects of targeted interventions on the internal processing of the system. Conformity, on the other hand, does not provide such causal explanations or constrain the internal causal processing of a system. To illustrate, recall the example described above about classifying images of cats and dogs. It would be possible to distinguish between the two sets of rules described there by introducing new data points—such as a dog with floppy ears and slitted pupils—and studying the system's behavior. Yet, these rules still describe only the behavior of the system, and do not provide a causal explanation of why those rules are an apt description. In particular, such a description does not identify or localize particular internal causally-relevant structures, nor facilitate interventions to change the observed behavior.

---

[2] It is worth noting that even this metaphysical notion of tacit knowledge has a subject-relative dimension to it. Whether or not a causally-relevant structure in the internal processing of an LLM can be given a particular semantic description (e.g., as an instance of a particular rule) depends on the concepts used in that description: while some intricate causal structures might not be describable in our day-to-day conceptual repertoire, they might nevertheless afford a semantic description in a more specialized (e.g. scientific or technical) vocabulary. Ultimately, determining which—if any—semantic description can be given to a causally-relevant structure will therefore depend in some way on an external observer who aims to explain the behavior of the system.

It should be noted that the notion of 'tacit knowledge' is historically controversial. First, while a causal-explanatory structure as specified by Davies' account of tacit knowledge might help explain the behavior of a speaker or LLM, it is controversial whether referring to this as 'knowledge' is appropriate at all (see e.g. Gascoigne and Thornton 2014; Davies 2015). Having knowledge is often closely associated with having justified true beliefs with a particular conceptual or propositional content. For instance, knowing the capital of France requires some concept of France and what it means to be a capital of a country. Yet, it might be argued that states of tacit knowledge typically do not have such conceptual content: a speaker might have tacit knowledge of a syntactic rule, but not possess the concepts to actually specify this rule. Moreover, states of tacit knowledge might not be considered beliefs insofar as a subject is often unaware of their tacit knowledge and states of tacit knowledge are mostly employed in specific circumstances—for instance, forming sentences—whereas beliefs can be used in a wider range of contexts (Evans 1985; Davies 2015). While I am aware of these debates, it is beyond the scope of the paper to address these concerns in detail. Nevertheless, I will adopt the term 'tacit knowledge' to stay in line with the precedent set by Davies as well as the relevant technical and theoretical literature in the AI community (e.g. Lam 2022; Meng et al. 2022). In doing so, I will focus on the causal-explanatory role tacit knowledge might play in the explanation of the behavior of LLMs, and bracket discussions about whether attribution of tacit knowledge warrants the attribution of stronger notions of knowledge.

Second, Quine raised an objection to Chomsky's account of tacit knowledge (1965), that is relevant to Davies' account as well. Chomsky suggested that human linguistic behavior could be explained through tacit knowledge, insofar as speakers of a language seem to have knowledge that guides their linguistic behavior, even if they are unable to verbalize the grammatical rules they use to structure a sentence. In response to this, Quine (1970) asked how we can distinguish between the many different sets of rules that adequately describe behavior, if no particular set of rules is explicitly represented by the language-user. In other words, how can we be sure that a given rule actually guides behavior if it is merely represented tacitly?

Taking into account these concerns, there are two reasons that Davies' account of tacit knowledge is particularly promising for explaining the (linguistic) behavior of LLMs. First, given that LLMs do not explicitly represent knowledge in the sense of allowing an observer to "read off" such knowledge, it seems reasonable to ask whether there are any implicit representations that guide their behavior. While Chomsky's account focuses on human linguistic behavior and is not directly applicable to LLMs, Davies' account targets neural networks and can thus be translated to LLMs with relatively minor modifications. Second, Davies' account allows for an answer to Quine's concern, insofar as it spells out clear constraints that should be met by a network in order to be attributed tacit knowledge of rules.

## 3.3. Davies' account of tacit knowledge

To specify what is required for attributing tacit knowledge, Davies introduces three constraints: *semantic description*, *causal systematicity*, and *syntactic structure*. To further characterize these constraints and thereby explain Davies' account of tacit knowledge, I will use a toy network with two outputs—"Paris" and "Berlin"—as an example. The inputs to this network are text prompts referring to particular tourist attractions in these two cities. For example, inputs to the system could be "The Brandenburg Gate is located in …", "The Sacré Coeur is located in the city of …", "To see the Eiffel Tower, I need to go to …". We assume that the network has been trained to successfully predict the next word for these prompts. Based on the behavior, it seems that the network has weak knowledge of the location of tourist attractions: it is able to correctly predict the output based on a particular input sentence. So what does it take for this network to be attributed tacit knowledge?

### 3.3.1 Semantic description

The first constraint for attributing tacit knowledge is *semantic description.* Davies argues that tacit knowledge can only be attributed to systems whose input and output states can be described semantically in the sense that they register or represent something. Davies does not spell out in detail what it means for an input or output state to have a semantic description, and I will not do so either.[3] As the inputs and outputs of LLMs are word sequences that can be considered to refer to something in a conventional sense, it seems reasonable to conclude that they meet this constraint.

Beyond ensuring that the input and output states of the system have semantic properties, semantic description might also reveal certain patterns of semantic similarity in the data. Recall the toy network described above. The inputs and outputs of this network have a particular semantic structure: there are two types of outputs, namely "Paris" and "Berlin". The inputs are statements about tourist attractions in these respective cities, and the network predicts in which of these cities the tourist attractions are. Then, there is a semantic pattern in the data insofar as inputs like "The Eiffel Tower is the main tourist attraction in …", "To see the Eiffel Tower, you have to travel to the city of …", etc. all refer to the same fact, namely that the Eiffel Tower is in Paris. Similarly, input statements referring to the Brandenburg Gate being in Berlin can be considered semantically similar as well.

This is of course a very simple example and semantic similarity is expected to be more complex in the actual inputs to real-world LLMs. Nevertheless, this example provides a simple way to describe a particular semantic pattern in the input-output data of the network. While semantic patterns by themselves do not indicate whether a network has tacit knowledge or not, Davies' second constraint is defined in relation to such semantic patterns in the data.

---

[3] For further discussion on this topic, see e.g. Dretske (1981), Millikan (1984).

**3.3.2 Causal systematicity**

The second and main constraint for attributing tacit knowledge is *causal systematicity*. Causal systematicity puts a constraint on the internal processing of the system: the internal processing in the network should mirror the semantic structure in the input-output pairs. For instance, there should be one causal process that mediates all input-output pairs referring to the Eiffel Tower, another one for all input-output pairs referring to the Brandenburg Gate, and so forth (see figure 2a). As Davies describes this constraint (1990, 88/89): "If we now consider several input-output pairs which exhibit a common pattern, we can ask whether those input-output transitions have a common causal explanation corresponding to the common pattern that they instantiate. Is there, within the physical system, a component mechanism or processor that operates as a causal common factor to mediate those several transitions? If there is a common causal explanation, then we can say that the process leading from those inputs to outputs is causally systematic, relative to that pattern."

In other words, the network should exhibit a pattern of causal common factors, which are structures in the network that mediate transitions from inputs to outputs for input-output pairs of a particular type, and as such serve as a common causal explanation for these transitions. This pattern of causal common factors should not just mirror any pattern in the data, but should correspond to the semantic patterns in the input-output data, for example the fact that the Eiffel Tower is in Paris. If such a pattern of causal common factors exists in the network, the network can be said to be causally systematic with regards to the semantic description of the input and output states.

Contrast this with a different underlying causal-explanatory structure, namely a network that has memorized all individual input-output pairs—a structure that is commonly called overfitting. In this case, there is an independent causal process that mediates each individual input-output pair (see figure 2b). As such, the network processes each input as being independent from all other inputs, and does not learn that there are semantic similarities within the data. Given the semantic structure of language, this pattern of memorization does not reflect the correct semantic patterns and therefore the network cannot be said to have tacit knowledge of semantic rules.

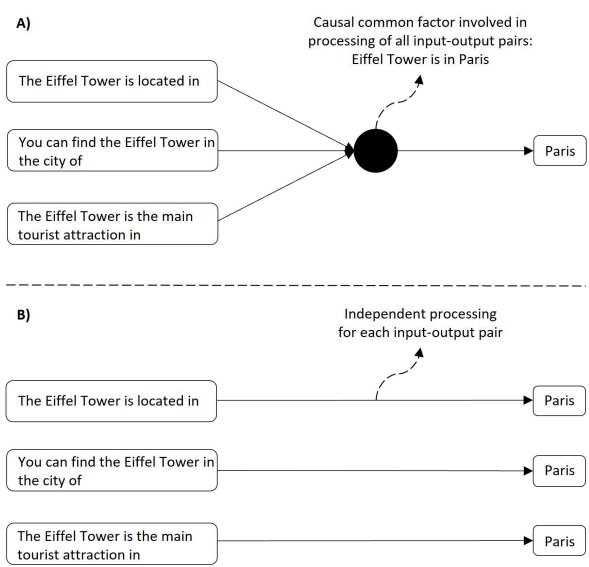

Figure 2: Examples of internal processing. A) shows causally systematic processing, with three semantically similar inputs, and a causal common factor that is involved in the processing for all three input-output transitions. B) shows an example of a network that memorizes input-output pairs: there is a one-to-one mapping between individual inputs and outputs.

As such, the main requirement for attributing tacit knowledge is that the internal structure of the network is causally systematic with regards to semantic patterns in the input-output data: there should be a pattern of causal common factors that mediates transitions from inputs to outputs in semantically similar cases. Going back to the toy example, this would mean that all transitions from inputs referring to the Eiffel Tower to outputs referring to Paris could be explained by one causal common factor, with another causal common factor mediating transitions for input-output pairs referring to the Brandenburg Gate.

      Briefly returning to the constraint of semantic description, note that causal common factors might, but need not, themselves have semantic descriptions. Insofar as the pattern of causal common factors mirrors the semantic structure of the data, Davies might have assumed that these internal factors are interpretable in the sense of being meaningful to human observers. Yet, given the black-box nature of connectionist networks, it is as of yet unknown whether these systems exhibit any causal common factors at all, and if so, whether these causal common factors are interpretable in the sense of allowing for semantic description. At the same time, it seems likely that the causal structure that ensures that an LLM performs well at predicting words of a particular natural language, acquired through extensive training, does so because that structure reflects something about the semantic structure of that language (see e.g. recent mechanistic interpretability work by Templeton and colleagues (2024)). Although I will follow Davies in not requiring semantic

descriptions of the internal structures for the attribution of tacit knowledge, determining if potential causal common factors can be given a semantic description seems like a particularly promising direction of future research.

The requirement of causal systematicity also allows for a response to Quine's concern about tacit knowledge. Consider the examples in figure 2, one adhering to the requirement of causal systematicity and one consisting of mere memorization. Although these systems might not be distinguishable based on behavior, experiments—in particular using interventions—can be used to identify the underlying causal structure (Davies 1987; Evans 1985). That is, changing the value of a causal common factor in a causally systematic system should lead to different outputs for all inputs of a particular type (figure 3a). Interventions on the model internals of a network with one-to-one mappings, on the other hand, would have localized effects (figure 3b). Therefore, interventions are one way to distinguish between different causal-explanatory structures and to verify whether a network meets the constraint of causal systematicity (see also section 5).

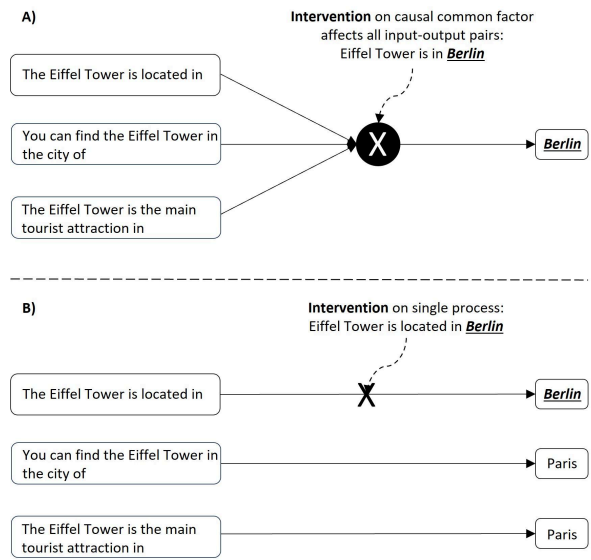

Figure 3: These structures are the same as in figure 2, but illustrate the effect of interventions. A) shows the effect of intervening on a causal common factor. Since a causal common factor is implicated in all transitions from inputs to outputs of a particular type, intervention on this causal common factor will affect all corresponding outputs. B) In the case of memorization, intervening on a single input-output pair will only affect the output for that particular input, and not affect any other input-output pairs.

### 3.3.3 Syntactic structure

The last constraint—syntactic structure[4]—serves as a bridge connecting semantic description with causal systematicity. Causal systematicity requires that all inputs that have a similar meaning—i.e. the same semantic description—should be processed by the same causal common factor (figure 2). To ensure that all semantically similar inputs are recognized and processed in a similar way, Davies argues that there should be a shared property within input representations that determines their causal processing. More precisely, semantically similar inputs should have some shared property that 1) allows the network to recognize all inputs of a particular (semantic) type and 2) determines the causal processing for these inputs. For example, there should be one unit or pattern of activation at the input layer that is active whenever there is an input about the Eiffel Tower, and another unit or pattern of activation that is active for inputs about the Brandenburg Gate. Whenever a new input contains one of the respective representations, the corresponding causal common factor is engaged. This ensures that all semantically similar inputs are processed by the same causal common factor.

---

[4] Note that Davies' use of 'syntax' is different than in linguistics and NLP, where 'syntax' generally refers to grammatical rules. In Davies' work, 'syntax' specifically refers to the physical properties of the input that determine the causal consequences.

## 3.4. Davies' objection

According to Davies, all three constraints—semantic description, causal systematicity, and syntactic structure—are necessary for attributing tacit knowledge to a connectionist neural network. Yet, Davies argues that connectionist neural networks with distributed representation do not meet the constraint of syntactic structure, as they do not have shared properties across representations of semantically similar inputs, due to the so-called *dimension shift* (Smolensky 1988). Whereas symbols in symbolic systems are the same regardless of context, neural networks represent their inputs as patterns of distributed activation. These patterns might be slightly different depending on the context of the input.

For example, while both "drinking coffee" and "a cup of coffee" contain the word 'coffee', their representations in the network are likely to differ given the different context. As such, there is no single pattern of activation—i.e. no syntactic property—that is shared by all inputs that contain the word 'coffee'. Because syntactic structure determines the causal processing of the inputs, the lack of a shared property means that different causal processes may be engaged and there is no single causal common factor engaged in all these input-output transitions. In other words, without syntactic structure, there cannot be causal systematicity and the network cannot be attributed tacit knowledge.

**4. A Reply: Architectural Innovations in LLMs**

In the remainder of the paper, I endorse the constraints of semantic description and causal systematicity as defined by Davies. However, Davies' constraint on syntactic structure might be too stringent for LLMs. In light of architectural differences between LLMs and earlier connectionist systems, I argue that LLMs meet a weaker version of the syntactic structure constraint that does not require identical properties across input representations, but that nevertheless suffices to attribute tacit knowledge. This weakened criterion for syntactic structure requires that the model maps semantically similar inputs to proximate regions in the input vector space through the embedding layer. This mapping allows the network to recognize and process inputs via a shared causal mechanism. Unlike Davies' original constraint, which requires an identical shared property in the input representation, this weaker constraint acknowledges the role of the embedding layers in recognizing semantically similar inputs and ensuring similar processing.

Let me elaborate on the weaker constraint and why it can be considered to have the same function, for the purpose of attributing tacit knowledge, as the more stringent constraint proposed by Davies. Presumably, being able to recognize semantically similar inputs is one of the reasons that transformer-based LLMs have been so successful. Whereas connectionist networks generally consist of architecturally (almost) identical layers—except for the trained weights and number of nodes—transformer-based models consist of specialized layers that take different roles in the processing of input sequences, such as the attention and feedforward layer (section 2). For the current discussion, I focus on the *embedding layer,* which computes a *word embedding* for each word in the input.

A word embedding is a representation of a word in the input that not only contains information about the word itself, but also about the context and the relationship to other words in the vocabulary. An important characteristic of a word embedding is that similar words are represented by similar embedding vectors, and dissimilar words are represented by more dissimilar embedding vectors, i.e. whose values are less alike (Mikolov et al. 2013; Grand et al. 2022). As such, embeddings of words with a similar meaning are not necessarily identical, but will be closer together in vector space than embeddings of words with more dissimilar meanings. This way of looking at the meaning of words, as given by their context and co-occurrence with other words, originates in the *distributional hypothesis*, which states that words that occur in a similar context will tend to have similar meanings (Firth 1957; Harris 1954).

As such, the embedding layer in LLMs acts as a kind of *semantic categorizer* which groups similar inputs together in a high-dimensional vector space. For example, it has been shown that the embedding vectors at various stages of processing reflect semantic and topical properties of the input, such as color similarity (Abdou et al. 2021), linguistic structure (Chang, Tu, and Bergen 2022), and word meaning (Grand et al. 2022). This means that the embedding layer allows the network to recognize semantically similar inputs and to make use of this information to predict the most likely next word.

Moreover, the fact that semantically similar inputs are close together in the high-dimensional vector space also guides the network towards similar processing pathways. While it is difficult to confirm empirically that these pathways are identical and research on how LLMs categorize and process their inputs is ongoing (e.g. Templeton et al. 2024), recent indirect evidence from intervention studies supports the conclusion that similar inputs are processed in similar ways (De Cao, Aziz, and Titov 2021; Meng et al. 2022). In particular, as will be illustrated in section 5, intervening on the network to change the prediction for "Paris is the capital of France" to "Paris is the capital of Germany" was shown to generalize to semantically similar inputs, e.g. paraphrases and other variations of the inputs. This suggests that the same internal processing is involved for these predictions, even in the absence of identical input vectors. By capturing the essence of syntactic structure through proximity in the embedding space, LLMs can thus achieve a form of causal systematicity without requiring identical activation patterns for semantically similar inputs.

In summary, Davies' requirement of identical syntactic properties for semantically similar inputs seems to be too strong in the context of LLMs. While transformer-based LLMs do not have identical syntactic representations at the input layers, the weaker criterion acknowledges the role of the embedding layer in achieving semantic clustering and consistent causal processing. This adjustment enables the attribution of tacit knowledge to LLMs by satisfying a more flexible yet functionally equivalent version of the syntactic structure constraint.

**5. Do LLMs have Tacit Knowledge?**

In the previous section, I have shown that transformer-based LLMs meet a weaker version of the syntactic structure constraint, which still allows for the attribution of tacit knowledge. The next step is to verify whether LLMs meet the constraint of causal systematicity. In this section, I analyze a recent example from the technical literature (Meng et al. 2022) and show that the representations of factual associations identified in this work could be considered causal common factors in Davies' sense, providing compelling—albeit preliminary—evidence that at least some LLMs meet the constraints for tacit knowledge.

**5.1 An example**

In their paper, Meng and colleagues investigate factual associations in LLMs, specifically where and how facts are stored in the model internals. Take the example of an LLM that correctly predicts "Paris" in response to an input like "The Eiffel Tower is located in", "Berlin" to "The Brandenburg Gate is located in the city of", etc. While it is often suggested that such a network "knows" the location of particular tourist attractions purely based on its behavior, Meng and colleagues investigate where the network stores such facts and how potential representations are used in the processing of the network.

Meng and colleagues approach this problem in two steps: first, localizing potential representations of facts, and then, verifying that these play a causal role in the observed behavior. To localize representations of facts in the network, Meng and colleagues use a method called *causal tracing,* which determines which activations, for example at which layer, are most important for a particular prediction. To test this, the activations of the network are corrupted such that the network no longer predicts the correct outcome. Then, activations are iteratively restored to the original value, until the network returns the correct prediction. Using this method, activations in the MLP layers are found to play an important role in representing factual associations.

Based on these results, Meng et al. suggest that facts are stored in certain MLP modules in the middle layers of the network. Moreover, in line with Geva and colleagues (2021), they suggest that MLP modules might function as so-called *key-value pairs*, which take inputs that represent a particular input and return outputs that reflect memorized properties of that input. For example, given an input "the Space Needle is in downtown", "Space Needle" might be viewed as a key, which prompts the MLP module to recall a particular value which represents properties of the Space Needle, such as its location. This information is accumulated over multiple layers and finally leads to a particular output at the last layer.

In the second part of their study, Meng and colleagues verify whether MLP modules function as key-value pairs and, if so, whether they can be updated to change the model's behavior. To do this, they introduce a method called *rank-one model editing* (ROME) which identifies and applies edits to an LLM's weights, based on two assumptions. First, if the MLP module functions as a key-value pair, this key-value pair can presumably be updated to include new information. Second, if the key-value pair plays a causal role in the model's behavior, updating the stored information should be reflected in the model's output. As such, the goal is to calculate a new key-value pair that encodes a particular updated factual association, e.g. "The Eiffel Tower is in Rome". The existing key-value pair is then replaced with this new key-value pair (see figure 4), after which the output of the network is evaluated for various prompts. The edit is successful if the output of the network changes to the newly inserted fact.

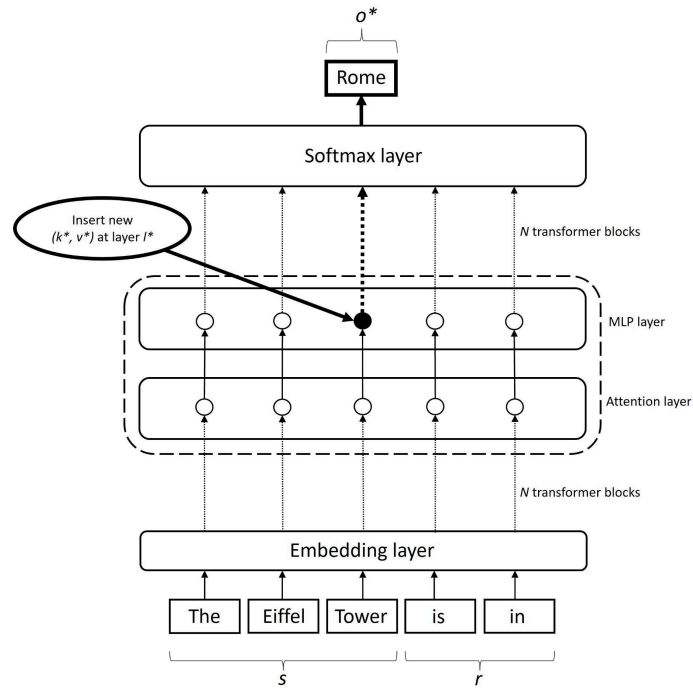

Figure 4: This figure shows an example of an intervention that is optimized to associate "The Eiffel Tower is in" with "Rome". The target is an MLP module in an intermediate layer of the network. In this layer, the key-value pair *(k, v)* at the last token of the subject (in this case, "Tower") is replaced with an updated key-value pair *(k\*, v\*)* that is optimized for the new association. Redrawn and adapted from Meng et al. (2022).

As mentioned, Meng et al.'s goal is to update what they call the "factual knowledge" of the system, in order to change its behavior. More precisely, what Meng et al. consider to be knowledge is the ability to generalize across contexts. This is contrasted with the network merely memorizing sequences of words from the input data and regurgitating these in response to a prompt. If an edit successfully targets this kind of knowledge, the updated fact should generalize: the network not only returns the updated fact for the exact input prompt, but also other semantically similar paraphrases. Yet, while Meng et al. operationalize knowledge as generalization, my goal is to determine if this can be considered a more robust kind of knowledge, namely tacit knowledge in Davies' sense. In the following sections, I show that MLP key-value pairs could be considered causal common factors, providing preliminary evidence that some LLMs can meet the constraint of causal systematicity and might thus have tacit knowledge.

### 5.2 MLP key-value pairs as causal common factors?

Imagine a causally systematic network with a pattern of causal common factors that reflects semantic similarity in the data. What would be the expected pattern of behavior in response to an intervention? In this network, an intervention successfully targeting a causal common factor would have a generalized effect. That is, intervening on a causal common factor that is implicated in a causal process $X$ would be expected to change all outputs that depend on $X$ (see figure 3). For example, changing a causal common factor that represents information about the capital of France to state "The capital of France is Berlin" would be expected to have an effect on all outputs referring to the capital of France. Moreover, this intervention would not affect transitions for other input-output pairs, i.e. the model does not suddenly predict that the capital of all countries is Berlin. In other words, a common causal factor should be involved in all transitions of input-output pairs of a given type, and not be involved in any other input-output transitions.

Based on two metrics—*generalization* to measure whether the update generalizes to semantically equivalent inputs, and *specificity* to measure whether the update does not affect unrelated input-output pairs—Meng and colleagues report that their edits update predictions both for the target input-output pair and for semantically similar input-output pairs. Although their results do not suggest that the network is causally systematic for all inputs—in which case generalization and specificity would both be 100%—the network largely seems to follow the expected pattern for causal systematicity. Even though Meng and colleagues only applied edits to single input-output pairs, generalization is high. This means that interventions on a specific fact update the predictions for semantically similar input-output pairs as well, for example paraphrases of the original input. Moreover, they report relatively high specificity suggesting that representations that concern different concepts are relatively disentangled. Taken together, this suggests that the target of intervention—the MLP module—does in fact function as a kind of causal common factor.

**5.3 Implications and reasons for caution**

Meng et al.'s results provide compelling preliminary evidence that at least some LLMs meet Davies' constraints, and can therefore be attributed tacit knowledge. LLMs meet the first constraint, semantic description, as their inputs and outputs consist of word sequences. The second constraint, causal systematicity, is also met, insofar as the LLM investigated by Meng and colleagues exhibits behavior characteristic of causally systematic systems in response to interventions. Finally, while Davies argues that connectionist neural networks do not meet the third constraint of syntactic structure, I have shown that LLMs meet a weaker version of this constraint, as word embeddings allow them to recognize semantically similar inputs. Moreover, the fact that interventions on MLP key-value pairs update both the target input-output pair as well as semantically similar paraphrases provides further empirical support for the claim that LLMs not only recognize semantically similar inputs, but also process these in a similar way.

Despite the promising results, there are some reasons for caution. First, although the reported specificity and generalization are high, they are not perfect. As such, updates might affect unrelated input-output pairs or fail to generalize. Possible explanations are that some updates do not target a causal common factor, or that causal common factors are not perfectly delineated in neural networks with distributed representation. The latter is of particular concern, as networks with distributed representation often exhibit *polysemanticity* or *superposition*, meaning that the same nodes in a network are involved in different decision processes (Van Gelder 1992; Elhage et al. 2022). Interventions might then affect multiple predictions, or even lead to catastrophic forgetting of previously learned associations. Further empirical research should determine to what extent this is a problem for interventions in practice.

Second, the work by Meng and colleagues (2022) relies heavily on interventions to locate and identify representations of tacit knowledge. Such interventions have recently been criticized, however. Specifically, while replicating the results reported by Meng and colleagues, Hase and colleagues (2023) found that interventions in different locations have a similar efficacy to the ones applied in the original study. From this, Hase and colleagues conclude that while causal tracing might localize a fact to a particular MLP key-value pair, this might not always be the best target for editing this fact. In the context of tacit knowledge, this means that causal tracing might not reliably localize the relevant causal common factors in the network.

More research is thus needed to investigate how causal common factors are represented in LLMs. Indeed, while Meng et al. seem to assume that facts are stored in a single layer—by focusing on layers during causal tracing—causal common factors, and thus tacit knowledge, might be represented over multiple layers. Neural networks often exhibit redundancy, meaning that various layers perform a similar task. As such, causal tracing might only partially characterize the relevant causal common factors (if there are any). While the work by Meng and colleagues provides preliminary evidence for the claim that neural networks can and do at least sometimes acquire tacit knowledge, the notion of 'causal common factors' and how these factors are represented in the relevant networks deserves further investigation.

## 6. Conclusion

The contributions of this paper are twofold. As a methodological contribution, I argued that we can take inspiration from Davies' account of tacit knowledge to conceptualize semantic knowledge in LLMs. More precisely, Davies' account provides clear criteria that should be met in order to attribute tacit knowledge to a particular system. While Davies himself argued that connectionist systems cannot meet the constraint of syntactic structure, I argued that this constraint can be appropriately weakened to acknowledge the role of the embedding layer in current LLMs. Armed with this weakened constraint, LLMs can in fact be attributed tacit knowledge, provided that they meet the other constraints. As an empirical contribution, I evaluated the recent work by Meng and colleagues (2022) to argue that there is compelling preliminary evidence to suggest that some LLMs actually acquire tacit knowledge in this sense. While this evidence is as of yet preliminary, tacit knowledge could thus be a promising tool to guide further research into the internal causal processing of LLMs.

      Notably, this paper fits into a broader literature suggesting that LLMs represent something akin to knowledge in the model internals (Hase et al. 2021; Li et al. 2023; Pavlick 2023). Provided this is supported by future work, this would have promising implications for explainable AI. In particular, the identification of such knowledge structures would not only provide a novel way to explain how these systems work, by

identifying internal common causal factors, but also to improve the performance of these systems, by applying interventions to change the internal knowledge representation so as to e.g. counteract misinformation, hallucination, and bias.

In this context, a promising direction of future research is the notion of interventions. Interventions are often used to investigate the causal processing within neural networks and to update the behavior of the network for particular inputs. In both cases, interventions are given a causal interpretation. It is not clear, however, to what extent such a causal interpretation is warranted for interventions as currently applied in machine learning. In particular, a concern is that interventions might have unintended results, for example affecting more predictions than intended. One way of approaching this problem could be to use existing frameworks like interventionism that lay out criteria for causal interventions (Woodward 2003).

In general, showing that LLMs have tacit knowledge would provide further insight into what LLMs can learn from pure-text data. Whereas it has been argued that current training methods for LLMs are insufficient for acquiring meaning (Bender and Koller 2020) or social aspects of language (Mahowald et al. 2024), tacit knowledge provides one way to conceptualize knowledge of semantic rules in LLMs. Since Meng and colleagues only analyze an early type of LLM, GPT-J, the question remains whether more recent LLMs also acquire tacit knowledge. Given the improvements in performance, however, this does not seem like an unreasonable assumption.

# References

Abdou, Mostafa, Artur Kulmizev, Daniel Hershcovich, Stella Frank, Ellie Pavlick, and Anders Søgaard. 2021. "Can Language Models Encode Perceptual Structure Without Grounding? A Case Study in Color." arXiv. https://doi.org/10.48550/arXiv.2109.06129.

Bahdanau, Dzmitry, Kyung Hyun Cho, and Yoshua Bengio. 2015. "Neural Machine Translation by Jointly Learning to Align and Translate." In *Proceedings of the 3rd International Conference on Learning Representations (ICLR 2015)*. San Diego, United States.

Bender, Emily M., and Alexander Koller. 2020. "Climbing towards NLU: On Meaning, Form, and Understanding in the Age of Data." In *Proceedings of the 58th Annual Meeting of the Association for Computational Linguistics*, 5185–98. Online: Association for Computational Linguistics. https://doi.org/10.18653/v1/2020.acl-main.463.

Block, Ned. 1981. "Psychologism and Behaviorism." *The Philosophical Review* 90 (1): 5–43. https://doi.org/10.2307/2184371.

Buckner, Cameron J. 2023. *From Deep Learning to Rational Machines: What the History of Philosophy Can Teach Us about the Future of Artificial Intelligence*. Oxford University Press.

Burrell, Jenna. 2016. "How the Machine 'Thinks': Understanding Opacity in Machine Learning Algorithms." *Big Data & Society* 3 (1): 205395171562251. https://doi.org/10.1177/2053951715622512.

Chang, Tyler, Zhuowen Tu, and Benjamin Bergen. 2022. "The Geometry of Multilingual Language Model Representations." In *Proceedings of the 2022 Conference on Empirical Methods in Natural Language Processing*, 119–36. Abu Dhabi, United Arab Emirates: Association for Computational Linguistics. https://doi.org/10.18653/v1/2022.emnlp-main.9.

Chomsky, Noam. 1965. *Aspects of the Theory of Syntax*. Cambridge, MA, USA: MIT Press.

Clark, Andy. 1991. "Systematicity, Structured Representations and Cognitive Architecture: A Reply to Fodor and Pylyshyn." In *Connectionism and the Philosophy of Mind*, edited by Terence Horgan and John Tienson, 198–218. Dordrecht: Springer Netherlands. https://doi.org/10.1007/978-94-011-3524-5_9.

Creel, Kathleen A. 2020. "Transparency in Complex Computational Systems." *Philosophy of Science* 87 (4): 568–89. https://doi.org/10.1086/709729.

Davies, Martin. 1987. "Tacit Knowledge and Semantic Theory: Can a Five Per Cent Difference Matter?" *Mind* 96 (384): 441–62.

Davies, Martin. 1990. "Knowledge of Rules in Connectionist Networks." *Intellectica. Revue de l'Association Pour La Recherche Cognitive* 9 (1): 81–126. https://doi.org/10.3406/intel.1990.881.

Davies, Martin. 2015. "Knowledge (Explicit, Implicit and Tacit): Philosophical Aspects." In *International Encyclopedia of the Social & Behavioral Sciences*, 74–90. Elsevier. https://doi.org/10.1016/B978-0-08-097086-8.63043-X.

De Cao, Nicola, Wilker Aziz, and Ivan Titov. 2021. "Editing Factual Knowledge in Language Models." arXiv. https://doi.org/10.48550/arXiv.2104.08164.

Dretske, Fred I. 1981. *Knowledge and the Flow of Information*. Cambridge, Massachusetts: MIT Press.

Elhage, Nelson, Tristan Hume, Catherine Olsson, Nicholas Schiefer, Tom Henighan, Shauna Kravec, Zac Hatfield-Dodds, et al. 2022. "Toy Models of Superposition." arXiv. https://doi.org/10.48550/arXiv.2209.10652.

Evans, Gareth. 1985. *Semantic Theory and Tacit Knowledge, in Collected Papers*. Oxford [Oxfordshire] : New York: Clarendon Press; Oxford University Press.

Firth, John Rupert. 1957. "A Synopsis of Linguistic Theory, 1930-1955." In *Studies in Linguistic Analysis*, edited by John Rupert Firth, 1–32. Oxford: Blackwell.

Fodor, Jerry A., and Zenon W. Pylyshyn. 1988. "Connectionism and Cognitive Architecture: A Critical Analysis." *Cognition* 28 (1): 3–71. https://doi.org/10.1016/0010-0277(88)90031-5.

Gascoigne, Neil, and Tim Thornton. 2014. *Tacit Knowledge*. London: Routledge. https://doi.org/10.4324/9781315729886.

Geva, Mor, Roei Schuster, Jonathan Berant, and Omer Levy. 2021. "Transformer Feed-Forward Layers Are Key-Value Memories." arXiv. https://doi.org/10.48550/arXiv.2012.14913.

Grand, Gabriel, Idan Asher Blank, Francisco Pereira, and Evelina Fedorenko. 2022. "Semantic Projection Recovers Rich Human Knowledge of Multiple Object Features from Word Embeddings." *Nature Human Behaviour* 6 (7): 975–87. https://doi.org/10.1038/s41562-022-01316-8.

Graves, Alex. 2012. "Long Short-Term Memory." In *Supervised Sequence Labelling with Recurrent Neural Networks*, edited by Alex Graves, 37–45. Studies in Computational Intelligence. Berlin, Heidelberg: Springer. https://doi.org/10.1007/978-3-642-24797-2_4.

Harris, Zellig S. 1954. "Distributional Structure." *WORD* 10 (2–3): 146–62. https://doi.org/10.1080/00437956.1954.11659520.

Hase, Peter, Mohit Bansal, Been Kim, and Asma Ghandeharioun. 2023. "Does Localization Inform Editing? Surprising Differences in Causality-Based Localization vs. Knowledge Editing in Language Models." *Advances in Neural Information Processing Systems* 36:17643–68.

Hase, Peter, Mona Diab, Asli Celikyilmaz, Xian Li, Zornitsa Kozareva, Veselin Stoyanov, Mohit Bansal, and Srinivasan Iyer. 2021. "Do Language Models Have Beliefs? Methods for Detecting, Updating, and Visualizing Model Beliefs." ArXiv. https://doi.org/10.48550/arXiv.2111.13654.

Hinton, G. E., J. L. McClelland, and D. E. Rumelhart. 1986. "Distributed Representations." In *Parallel Distributed Processing: Explorations in the Microstructure of Cognition, Vol. 1: Foundations*, 77–109. Cambridge, MA, USA: MIT Press.

Kripke, Saul A. 1982. *Wittgenstein on Rules and Private Language: An Elementary Exposition*. Harvard University Press.

Lam, Nardi. 2022. "Explanations in AI as Claims of Tacit Knowledge." *Minds and Machines*. https://doi.org/10.1007/s11023-021-09588-1.

Lapuschkin, Sebastian, Stephan Wäldchen, Alexander Binder, Grégoire Montavon,

Wojciech Samek, and Klaus-Robert Müller. 2019. "Unmasking Clever Hans Predictors and Assessing What Machines Really Learn." *Nature Communications* 10 (1): 1096. https://doi.org/10.1038/s41467-019-08987-4.

Lenat, Douglas B., and R. V. Guha. 1989. *Building Large Knowledge-Based Systems; Representation and Inference in the Cyc Project*. 1st ed. USA: Addison-Wesley Longman Publishing Co., Inc.

Li, Kenneth, Aspen K. Hopkins, David Bau, Fernanda Viégas, Hanspeter Pfister, and Martin Wattenberg. 2023. "Emergent World Representations: Exploring a Sequence Model Trained on a Synthetic Task." In *Proceedings of the 11th International Conference on Learning Representations (ICLR 2023)*. ICLR.

Linzen, Tal, and Marco Baroni. 2021. "Syntactic Structure from Deep Learning." *Annual Review of Linguistics* 7 (1): 195–212. https://doi.org/10.1146/annurev-linguistics-032020-051035.

Mahowald, Kyle, Anna A. Ivanova, Idan A. Blank, Nancy Kanwisher, Joshua B. Tenenbaum, and Evelina Fedorenko. 2024. "Dissociating Language and Thought in Large Language Models." *Trends in Cognitive Sciences* 28 (6): 517–40. https://doi.org/10.1016/j.tics.2024.01.011.

Meng, Kevin, David Bau, Alex Andonian, and Yonatan Belinkov. 2022. "Locating and Editing Factual Associations in GPT." In *Advances in Neural Information Processing Systems*, 35:17359–72.

Mikolov, Tomas, Ilya Sutskever, Kai Chen, Greg S Corrado, and Jeff Dean. 2013. "Distributed Representations of Words and Phrases and Their Compositionality." In *Advances in Neural Information Processing Systems*. Vol. 26.

Millikan, Ruth Garrett. 1984. *Language, Thought, and Other Biological Categories: New Foundations for Realism*. Cambridge, Massachusetts: MIT Press.

Nadkarni, Prakash M, Lucila Ohno-Machado, and Wendy W Chapman. 2011. "Natural Language Processing: An Introduction." *Journal of the American Medical Informatics Association* 18 (5): 544–51. https://doi.org/10.1136/amiajnl-2011-000464.

Olah, Chris, Nick Cammarata, Ludwig Schubert, Gabriel Goh, Michael Petrov, and Shan Carter. 2020. "Zoom In: An Introduction to Circuits." *Distill* 5 (3): e00024.001. https://doi.org/10.23915/distill.00024.001.

OpenAI. 2022. "ChatGPT: Optimizing Language Models for Dialogue." OpenAI. November 30, 2022. https://openai.com/blog/chatgpt/.

Pavlick, Ellie. 2023. "Symbols and Grounding in Large Language Models." *Philosophical Transactions of the Royal Society A: Mathematical, Physical and Engineering Sciences* 381 (2251): 20220041. https://doi.org/10.1098/rsta.2022.0041.

Piantadosi, Steven T., and Felix Hill. 2022. "Meaning without Reference in Large Language Models." arXiv. https://doi.org/10.48550/arXiv.2208.02957.

Quine, W.V. 1970. "Methodological Reflections on Current Linguistic Theory." *Synthese* 21:13. https://doi.org/10.1007/978-94-010-2557-7_14.

Shanahan, Murray. 2023. "Talking About Large Language Models." arXiv. https://doi.org/10.48550/arXiv.2212.03551.

Smolensky, Paul. 1988. "On the Proper Treatment of Connectionism." *Behavioral and*

*Brain Sciences* 11 (1): 1–23. https://doi.org/10.1017/S0140525X00052432.

Sukhbaatar, Sainbayar, Edouard Grave, Guillaume Lample, Herve Jegou, and Armand Joulin. 2019. "Augmenting Self-Attention with Persistent Memory." arXiv. https://doi.org/10.48550/arXiv.1907.01470.

Templeton, Adly, Tom Conerly, Jonathan Marcus, Jack Lindsey, Trenton Bricken, Brian Chen, Adam Pearce, et al. 2024. "Scaling Monosemanticity: Extracting Interpretable Features from Claude 3 Sonnet." *Transformer Circuits Thread*. https://transformer-circuits.pub/2024/scaling-monosemanticity/index.html.

Van Gelder, Tim. 1992. "Defining 'Distributed Representation.'" *Connection Science* 4 (3–4): 175–91. https://doi.org/10.1080/09540099208946614.

Vaswani, Ashish, Noam Shazeer, Niki Parmar, Jakob Uszkoreit, Llion Jones, Aidan N Gomez, Łukasz Kaiser, and Illia Polosukhin. 2017. "Attention Is All You Need." In *Advances in Neural Information Processing Systems*. Vol. 30.

Woodward, James. 2003. *Making Things Happen: A Theory of Causal Explanation*. Oxford University Press, USA.

Yildirim, Ilker, and L. A. Paul. 2024. "From Task Structures to World Models: What Do LLMs Know?" *Trends in Cognitive Sciences* 28 (5): 404–15. https://doi.org/10.1016/j.tics.2024.02.008.

Zednik, Carlos. 2018. "Will Machine Learning Yield Machine Intelligence?" In *Philosophy and Theory of Artificial Intelligence 2017*, edited by Vincent Müller, 44:225–27. Studies in Applied Philosophy, Epistemology and Rational Ethics. Springer International Publishing. https://doi.org/10.1007/978-3-319-96448-5_23.

Zednik, Carlos. 2021. "Solving the Black Box Problem: A Normative Framework for Explainable Artificial Intelligence." *Philosophy & Technology* 34 (2): 265–88. https://doi.org/10.1007/s13347-019-00382-7.

**Acknowledgements**: I am grateful to Carlos Zednik, Vincent Müller, Lambèr Royakkers, and Charles Rathkopf, as well as my colleagues at the Philosophy & Ethics group at Eindhoven University of Technology, for helpful discussions and comments. Versions of this paper were presented at the Early Career Researchers Workshop hosted by the Ruhr-Universität Bochum, the 49th meeting of the Society of Philosophy and Psychology at the University of Pittsburgh, and the 5th Conference on Philosophy of Artificial Intelligence hosted by the Universität Erlangen-Nuremberg. I would like to thank the participants for their helpful questions and comments.

**Funding information:** This research was funded by an EAISI startup grant "Cognitive models as surrogate models for explainable AI".

**Declarations:** None to declare.